%% file: main.tex
\documentclass[runningheads]{llncs}
\usepackage[T1]{fontenc}

\usepackage{graphicx}
\usepackage{tabularx}
\usepackage{placeins}
\usepackage{float}
\usepackage{colortbl} 

\usepackage{booktabs} 
\usepackage{xcolor} 
\usepackage{colortbl} 

\usepackage{color}
\definecolor{myred}{rgb}{.8,.0,.0}

\definecolor{mypink}{rgb}{1.0, 0.08, 0.58} 

\usepackage{booktabs}
\usepackage{caption}
\usepackage{subcaption}
\usepackage{framed,multirow}
\usepackage{amsmath}

\newcommand\blfootnote[1]{%
  \begingroup
  \renewcommand\thefootnote{}\footnote{#1}%
  \addtocounter{footnote}{-1}%
  \endgroup
}

\usepackage[breaklinks=true,bookmarks=false]{hyperref}

\ifdefined\DOUBLEBLIND

\else

\fi

\begin{document}


\title{Look What the Probes Dragged In!\\ Real-World Chest X-ray Shortcuts in MedCLIP}

\titlerunning{Real-World Shortcuts in MedCLIP}

\ifdefined\DOUBLEBLIND
    \author{***}
    \authorrunning{***}
    \institute{***}
\else
    \author{Nikolette Pedersen\inst{1,*} \and
    Regitze Sydendal\inst{1,*} \and
    Veronika Cheplygina\inst{1} \and \\
    Théo Sourget\inst{1} \\
    }
    \authorrunning{N. Pedersen et al.}
    \institute{
    Pattern Recognition Revisited Lab (PURRlab)\\
    IT University of Copenhagen, Denmark\\
    \email{\{nizp,resy,vech,tsou\}@itu.dk}\\
    }
\fi




\maketitle 

\begin{abstract}

\input{sec_00_abstract}

\end{abstract}

\section{Introduction} \label{sec:intro}
\input{sec_01_intro}

\section{Related Work} \label{sec:related}

\input{sec_02_related}

\section{Investigating MedCLIP's Shortcuts with Linear Probes} \label{sec:methods}
\input{sec_03_data}

\input{sec_03_methods}


\section{Results} \label{sec:results}
\input{sec_04_results}

\section{Discussion and Conclusions} \label{sec:conc}
\input{sec_09_lim_fut_con}

\begin{credits}
\subsubsection{\ackname}\label{sec:Acknowledgments}
\input{sec_10_ack}

\subsubsection{\discintname}\label{sec:Interests}
\input{sec_11_interests.tex}

\end{credits}

%
%
\newpage
\bibliographystyle{splncs04}
\bibliography{references}

\end{document}

%% file: sec_00_abstract.tex
Vision-language models, such as contrastive language-image pre-training (CLIP)-based approaches, have reached state-of-the-art (SOTA) results in medical artificial intelligence. However, recent work reveals that CLIP-based models remain vulnerable to shortcuts. We investigate how real-world shortcuts manifest across different layers of the medical CLIP-based model, MedCLIP, and its vision encoder, a frozen ResNet-50. We attach 17 linear classification probes to the intermediate layers of the ResNet-50 and train them on three different dataset configurations and targets: NIH-CXR14 (pneumothorax) and PadChest (cardiomegaly and pneumothorax). This setup allows us to observe model behaviour during evaluation using subgroup-based calibration and layer-wise confidence curves. We find that the final linear probes achieve a high AUROC but poor calibration in the models. The layer-wise confidence analyses suggest that shortcuts emerge at different depths. Patterns consistent with localised shortcuts, such as drains, appear at later layers, while patterns consistent with diffuse shortcuts, such as scanner-specific noise patterns, emerge earlier, aligning with previous work. Finally, we conduct a manual analysis of the images, which reveals data quality issues in both NIH-CXR14 and PadChest. Our findings underscore that even SOTA models remain vulnerable to shortcuts, and the need for high-quality and well-annotated datasets to draw solid conclusions. Code can be found on our GitHub: {\url{https://github.com/nikodice4/MedCLIP_shortcuts}}. \blfootnote{* Equal contribution}

%% file: sec_01_intro.tex
Vision-language models (VLMs), which leverage both images and medical reports, show high performance in medical AI, yet previous work raises questions about their fairness \cite{Luo_2024_CVPR}, and their tendency to rely on shortcuts \cite{geirhos_shortcut_2020} such as chest drains in pneumothorax classification~\cite{oaktree2020hidden}.

Recent studies \cite{sourget2025fairness,boland2025preventing} investigate these biases and how shortcuts arise during training, yielding results worth further investigation. For VLMs, Sourget et al.~\cite{sourget2025fairness} show that six contrastive language-image pre-training (CLIP)-based models perform better on chest X-rays with drains than without, indicating shortcut reliance. Three of the models, CXR-CLIP~\cite{tiu2022expert}, CheXzero~\cite{you2023cxr}, and MedCLIP \cite{wang-etal-2022-medclip}, produce predicted probabilities clustered around $0.5$ in zero-shot classification, despite CheXzero and MedCLIP achieving area under the receiver operating characteristic curve (AUROC) scores that are within typical ranges for these tasks. For convolutional neural networks (CNNs), Boland et al.~\cite{boland2025preventing} investigate layer-wise confidence using linear probes trained on synthetically biased data, and show that models trained on biased data tend to be overconfident \cite{boland2025preventing}. This makes MedCLIP worth exploring, since if a model relies on shortcuts, we would expect overconfidence, rather than near-random probabilities \cite{sourget2025fairness}. We explore MedCLIP in a linear probing setting, rather than a zero-shot classification setting.

Given these insights, we investigate MedCLIP's reliance on shortcuts across layers. We attach 17 linear classification probes to the intermediate building blocks in MedCLIP's vision encoder, a frozen ResNet-50, training them on three dataset configurations: pneumothorax classification on NIH-CXR14, and cardiomegaly and pneumothorax classification on PadChest. We then visualise the calibration and layer-wise confidence curves for each probe. Our contributions are as follows: i) implementing linear classification probes in MedCLIP and training the probes on real-world medical data, rather than curated biased data, ii) investigating MedCLIP's vision encoder (ResNet-50), revealing poor calibration and shortcut learning at different layers and iii) a manual analysis of the model predictions, highlighting the importance of data quality.

%% file: sec_02_related.tex
\indent Despite deep learning models reaching strong performance on chest X-ray classification~\cite{hosny2018artificial,shen2019artificial,tiu2022expert}, multiple works show that high-performance metrics alone do not reveal whether a model is learning clinically relevant features and could be impacted by shortcut learning~\cite{geirhos_shortcut_2020,vasquez2024detecting}. 
A typical shortcut in pneumothorax classification is the presence of a chest drain, which correlates with positive cases of pneumothorax. Oakden-Rayner et al.~\cite{oaktree2020hidden} find that a DenseNet-121 trained on NIH-CXR14 achieves an AUROC of $0.940$ on images with drains versus $0.770$ without, and Jim{\'e}nez-S{\'a}nchez et al.~\cite{10230572} find the same pattern on CheXpert and NIH-CXR14. Sourget et al.~\cite{sourget2025fairness} conduct a similar analysis for several CLIP-based models in a zero-shot setting and obtain similar findings, but also show the poor calibration of all models. 

Boland et al.~\cite{boland2025preventing} study how different kinds of shortcuts emerge at different layer depths by attaching linear probes to intermediate layers of CNNs trained on synthetically biased data. A diffuse shortcut is uniform noise spread across the image, and emerges in the earlier layers, while localised shortcuts, such as a red square placed at a fixed or random location in the images, appear in the later layers. They also introduce bias mitigation strategies targeting the intermediate layers using knowledge distillation from an unbiased teacher model.

Other works describe the quality and challenges of publicly available datasets and how these affect AI models. Rafferty and Rajan~\cite{rafferty-2025} identify four major dataset challenges: label quality and semantic noise from automated annotations, demographic and institutional biases, technical artefacts promoting shortcut learning, and limitations of evaluation practices. 
Testing across MIMIC-CXR, CheXpert, NIH-CXR14 and PadChest, they find that no architecture proves consistently robust to domain shift, establishing that dataset characteristics are the primary drivers of failure in chest radiography, rather than network design. Jim{\'e}nez-S{\'a}nchez et al.~\cite{jimenezsanchez2025picture} also discuss the current limitations of literature reviews focusing mostly on methods and less on datasets. They propose a living review to track datasets and their research artifacts, such as shortcuts, and discuss important considerations for dataset curation.

Building on the work of Boland et al.~\cite{boland2025preventing} and Sourget et al.~\cite{sourget2025fairness}, we aim to identify the depths at which real-world shortcuts emerge in MedCLIP and study its calibration with linear probes. We also perform a manual analysis highlighting data quality issues in large publicly available medical datasets.

%% file: sec_03_data.tex
\subsection{Datasets}
\subsubsection{NIH-CXR14}
This dataset was released in collaboration with the National Institutes of Health (NIH), containing $112,120$ chest X-ray images from $30,805$ unique patients, collected between 1992 and 2015, with 14 thoracic disease labels mined from reports using natural language processing techniques. We focus on pneumothorax, also known as collapsed lung, which occurs when air leaks into the space between the lung and the chest wall \cite{MCNAMARA2026197}.
\\
\indent To analyse drain-related shortcuts, we combine two supplementary annotation datasets. NEATX (Non-Expert Annotations of Tubes in X-rays) by \cite{10.1007/978-3-031-98688-8_10} provides manual chest drain annotations for $3,709$ positive samples of pneumothorax from NIH-CXR14. Since NEATX only contains positive cases of pneumothorax, we supplement it with automatically generated drain labels from \cite{sourget2025fairness}, who train a DenseNet-121 on the NEATX labels and apply it to the negative-pneumothorax cases in NIH-CXR14.
\input{tables/dataset_info}

\subsubsection{PadChest}
The second dataset is PadChest (Pathology
Detection in Chest radiographs), one of the largest open-source datasets containing $160,861$ chest X-ray images from $67,625$ patients collected at the San Juan Hospital of Alicante in Spain between 2009 and 2017, with multi-labelled annotated reports in Spanish. It covers five different projections: PA, L, AP, AP-horizontal, and COSTAL.
\\
\indent We use two target labels: cardiomegaly and pneumothorax. Cardiomegaly (enlarged heart)~\cite{amin2022cardiomegaly} was chosen for its higher class prevalence and sex balance across patients. Pneumothorax is heavily underrepresented at $0.4\%$, but allows for a comparison of the same disease across two datasets.
\\
\indent PadChest also includes three registered patient sexes: ``female'', ``male'' and ``other''. The 14 patients registered as ``other'' are included in the training set, but not represented in the sex subgroup plots.
\subsubsection{Preprocessing the Datasets}
For PadChest, we follow the preprocessing methodology of \cite{sourget_mask_2025}, and remove all vertical X-ray projections, keeping only the following projections: PA, AP, AP-horizontal. We also exclude rows where the disease label is ``suboptimal study'', ``exclude'' or ``Unchanged''. 
The NIH-CXR14 dataset already provides a train and test split. For PadChest, we split the dataset into a train (80\% of the data) and a test split (20\% of the data). Both train sets are then split again to use 20\% as validation data. All splits were done by patient identifier to avoid any data leakage. All information about the sizes of the datasets
and subgroups can be found in \autoref{tab:dataset_splits}.

%% file: tables/dataset_info.tex
\begin{table*}[t]
    \centering
    \caption{Complete overview of all datasets, stratified by their splits (training, validation, test) and diagnostic label, with further stratification by the subgroups (sex, scanner, drains) we investigate. Scanner types are abbreviated as IDC (ImagingDynamicsCompanyLtd) and PMS (PhilipsMedicalSystems) and positive and negative are abbreviated as Pos. and Neg.}
    \makebox[\textwidth]{%
    \resizebox{\textwidth}{!}{%
    \begin{tabular}{ll|rr|rr|rr|rr|rr|rr|rr|r|r}
    & & \multicolumn{2}{c|}{\textbf{Female Patients}} & \multicolumn{2}{c|}{\textbf{Male Patients}} & \multicolumn{2}{c|}{\textbf{Other Patients}} & \multicolumn{2}{c|}{\textbf{IDC}} & \multicolumn{2}{c|}{\textbf{PMS}} & \multicolumn{2}{c|}{\textbf{Drain}} & \multicolumn{2}{c|}{\textbf{No Drain}} & \multicolumn{1}{c|}{\textbf{Total}} & \multicolumn{1}{c}{\textbf{Pos. Ratio}} \\
    \textbf{Dataset} & \textbf{Split} & Neg. & Pos. & Neg. & Pos. & Neg. & Pos. & Neg. & Pos. & Neg. & Pos. & Neg. & Pos. & Neg. & Pos. & N & (\%) \\ \hline
    NIH-CXR14 & Train & 29,148 & 1,111 & 38,314 & 1,052 & - & - & - & - & - & - & - & - & - & - & 69,625 & 3.1 \\
     (Pneumothorax) & Val & 7,564 & 243 & 8,861 & 231 & - & - & - & - & - & - & - & - & - & - & 16,899 & 2.8 \\
     & Test & 9,483 & 1,231 & 13,448 & 1,434 & - & - & - & - & - & - & 10,494 & 1,692 & 12,437 & 973 & 25,596 & 10.4 \\
    \hline
    PadChest & Train & 30,862 & 3,518 & 32,030 & 2,705 & 6 & 1 & 31,649 & 3,034 & 31,249 & 3,190 & - & - & - & - & 69,122 & 9.0 \\
     (Cardiomegaly) & Val & 7,620 & 881 & 7,999 & 689 & 4 & 0 & 7,917 & 774 & 7,707 & 796 & - & - & - & - & 17,194 & 9.1 \\
     & Test & 9,716 & 1,044 & 10,183 & 822 & 3 & 0 & 10,061 & 909 & 9,843 & 957 & - & - & - & - & 21,770 & 8.6 \\
    \hline
    PadChest 
     & Train & 34,287 & 93 & 34,580 & 155 & 7 & 0 & 34,654 & 29 & 34,220 & 219 & - & - & - & - & 69,122 & 0.4 \\
     (Pneumothorax) & Val & 8,483 & 18 & 8,653 & 35 & 4 & 0 & 8,685 & 6 & 8,456 & 47 & - & - & - & - & 17,194 & 0.3 \\
     & Test & 10,734 & 26 & 10,949 & 56 & 3 & 0 & 10,966 & 4 & 10,722 & 78 & - & - & - & - & 21,770 & 0.4 \\
    \hline
    \end{tabular}
    }}
    \label{tab:dataset_splits}
\end{table*}

%% file: sec_03_methods.tex
\subsection{Linear Probe Setting in MedCLIP and Evaluation}
MedCLIP is a pre-trained VLM combining vision and language encoders, pre-trained on CheXpert, MIMIC-CXR, COVID, and RSNA Pneumonia \cite{wang-etal-2022-medclip}. The vision and text encoders map inputs to a shared embedding space, where the predicted cosine similarity between projected embeddings is aligned to a Unified Medical Language System (UMLS)-derived similarity matrix. Standard CLIP-based models require paired image-text data, which is scarce in medical settings, and treat semantically similar but unpaired samples as negatives. MedCLIP addresses both issues by decoupling image-text pairs using medical knowledge extracted via the UMLS, which enables training on unpaired data and dramatically increases the dataset size for training. 

To investigate how the classification confidence of MedCLIP's vision encoder develops throughout the layers, we follow the methodology of Boland et al.~\cite{boland2025preventing}. Each linear probe consists of an average pooling layer, followed by a single-linear fully connected neural network. They attach the probes to the intermediate layers of their CNN architectures.
\\
\indent We attach linear probes to the output of each building block of our frozen ResNet-50 architecture, and one final probe is placed at the last average pooling layer, providing the final predictions for the input. In total, we train 17 probes. To assess the calibration of the model, we use the predictions from the final probe, the main classification task being disease vs. no disease.

We train the attached linear probes three separate times, each time with a different dataset configuration. Only the weights of the linear probes are updated during training, as the frozen ResNet-50 captures representations from MedCLIP's pre-training. The training loss is the sum of the cross-entropy losses across all 17 linear probes (16 are intermediate, plus the final probe on the average-pooled output). Since the backbone is frozen and probes do not share weights, each probe receives gradients only from its own loss term. All probes were trained using the same hyperparameters: epochs: $100$, batch size: $32$, learning rate: $0.00001$. To mitigate overfitting, we monitor the summed validation loss after each epoch with early stopping after $15$ epochs without improvement.
\subsection{Calibration and Confidence Curves}
We plot the mean predicted probabilities from the final probe against the fraction of true positive cases of the given disease in each bin to produce our calibration curve. A perfectly calibrated model is one whose predicted probabilities match the true frequency of a given disease. This means, among all cases assigned a probability of $p$, a fraction $p$ are actually positive.
\\
\indent 
To quantify how classification confidence develops throughout the model, we follow the methodology of \cite{boland2025preventing}, who define the model's confidence, $C(X)$, as the deviation from the maximum uncertainty of $0.5$, where a higher value indicates a greater certainty in the prediction. \cite{boland2025preventing} find that the model becomes more overconfident when training on curated biased data than when training on clean data. As we are not dealing with synthetically biased data, we do not expect to see the exact same results. Following the taxonomy from \cite{10.1007/978-3-031-82007-6_11}, physical devices like chest drains are categorised as external shortcuts. We therefore hypothesise that we will see spikes in the later layers, as the chest drains act more like a localised shortcut than a diffuse one, since a physical device aligns with \cite{boland2025preventing}'s synthetic red squares rather than the synthetic noise. In contrast, the different X-ray machines are categorised as imaging shortcuts and can produce image noise, which we hypothesise will align with a diffuse shortcut. However, the IDC images also contain an ``R'' marker in the X-rays, which can resemble a localised shortcut.


%% file: sec_04_results.tex
\input{figures/auroc}

\input{figures/calibration_subfig}

\subsection{Good Overall AUROC Across Dataset Configurations}

\autoref{fig:auroc} shows the final probes' AUROC scores across all test set configurations. Pneumothorax on NIH-CXR14 achieves a global AUROC score of $0.839$, with a small gap of $0.028$ between drains ($0.840$) and no drains ($0.812$). There is no notable difference between the female and male patients' AUROC scores. Cardiomegaly on PadChest achieves the highest global AUROC score across all three classification tasks, at $0.905$. The subgroup IDC achieves the highest AUROC score across all dataset configurations and subgroups, reaching $0.917$, while its counterpart, PMS, reaches $0.893$. 
Pneumothorax on PadChest achieves a global AUROC score of $0.875$. PMS and IDC have AUROC scores of $0.801$ and $0.852$, respectively, with PMS having the lowest AUROC score across all subgroups. Further, we see large confidence intervals, which is due to the small number of positive cases in this dataset configuration. 

\subsection{Generally Poorly Calibrated Models}
\autoref{fig:calibration_subfig} shows calibration curves and confidence curves, with each column corresponding to one of the three dataset configurations. 
Pneumothorax on NIH-CXR14, split by whether a patient has a drain, shows the model is poorly calibrated and overconfident, though marginally better calibrated for the drain subgroup.
For cardiomegaly on PadChest, split by X-ray machine, we see that this is the best calibrated model, and it is also the dataset configuration with the most positive cases. Still, the model is overconfident, and we see no real difference between the subgroups. 
Pneumothorax on PadChest is the most miscalibrated model. The IDC subgroup is the worst calibrated. This is notably also the subgroup with the least positive cases. 
For the PadChest pneumothorax patient sex subgroup, it looks similar to its scanner machine counterpart, and there are no differences between the two subgroups.
Overall, every model is miscalibrated to some degree, and the number of positive cases for the dataset configurations visibly influences the curves.

\subsection{Confidence Curves Show Signs of Shortcuts}
For pneumothorax on NIH-CXR14, split by drain status, all four curves stay low and stable across the first 13 building blocks. The groups diverge from block 13, where the positive cases (drain in blue and no drain in orange) reach high confidence between $0.35$ and $0.41$, while the negative cases stay below $0.25$, however the green curve (drain) remains higher than the red curve (no drain). These findings align with \cite{boland2025preventing}, who found localised shortcuts spike in later layers. Cardiomegaly on PadChest, split by X-ray machine, shows all four curves remain low, spiking between building blocks 4 and 6 and again at building block 13. All four lines lie in the same range and yield a relatively low confidence, around $0.23$-$0.28$, therefore these spikes do not resemble the patterns of shortcuts. Pneumothorax on PadChest shows spikes in the earlier layers. With only 4 positive pneumothorax cases, the IDC curve (in blue) is very uncertain. All lines spike at building block 3 and afterwards remain stable in their trajectory. The IDC curve (in green) for the negative cases peaks at $0.42$, but all curves end at similar levels. These spikes do resemble the diffuse shortcuts of \cite{boland2025preventing}, which further aligns with previous research that also finds how different X-ray machines in an unbalanced setting can lead to shortcuts \cite{sourget2026datasetdiversitymetricsimpact}. However, the IDC X-rays also have an ``R'' marker in the corner, which on its own can resemble a localised shortcut. Therefore, the plots are not enough evidence to conclude whether the shortcuts are diffuse or localised.

%% file: figures/auroc.tex
\begin{figure}[t]
    \centering
    \includegraphics[width=\textwidth]{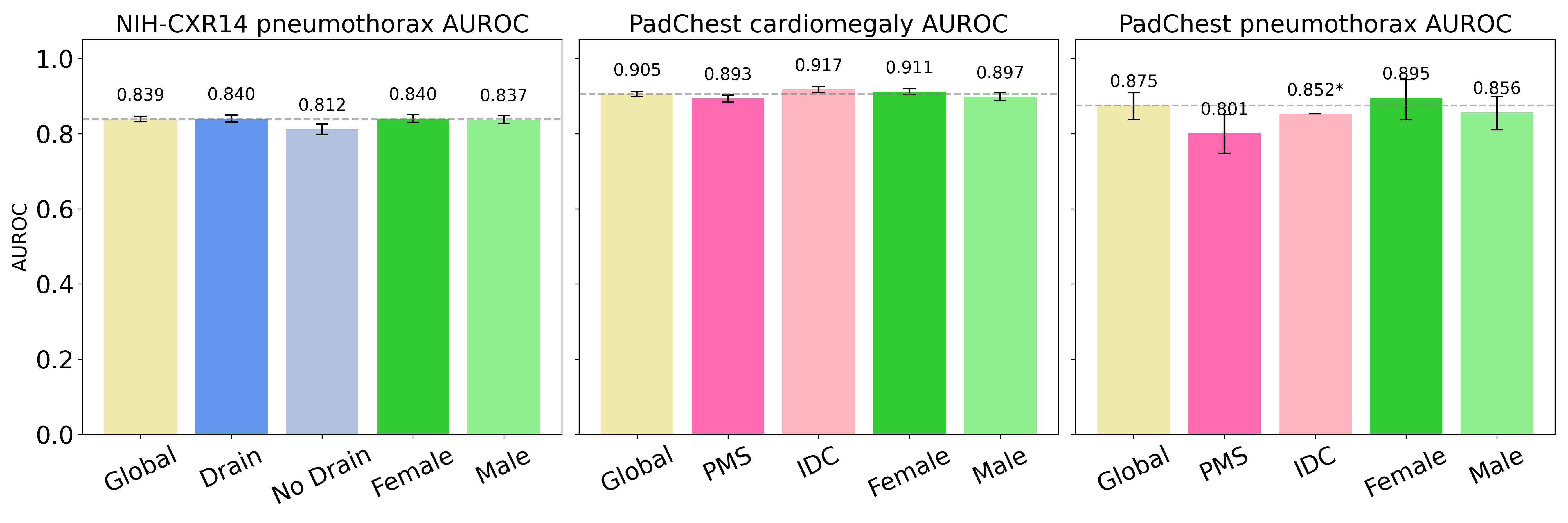}
    \caption{AUROC scores across all dataset configurations and their given subgroups, with the dashed lines indicating the global AUROC scores. The confidence intervals are displayed as black error bars. ``*'' means that there is no confidence interval as there are not enough positive samples for the bootstrapping.}
    \label{fig:auroc}
\end{figure}

%% file: figures/calibration_subfig.tex
\begin{figure}[t]
    \centering
    \includegraphics[width=\textwidth]{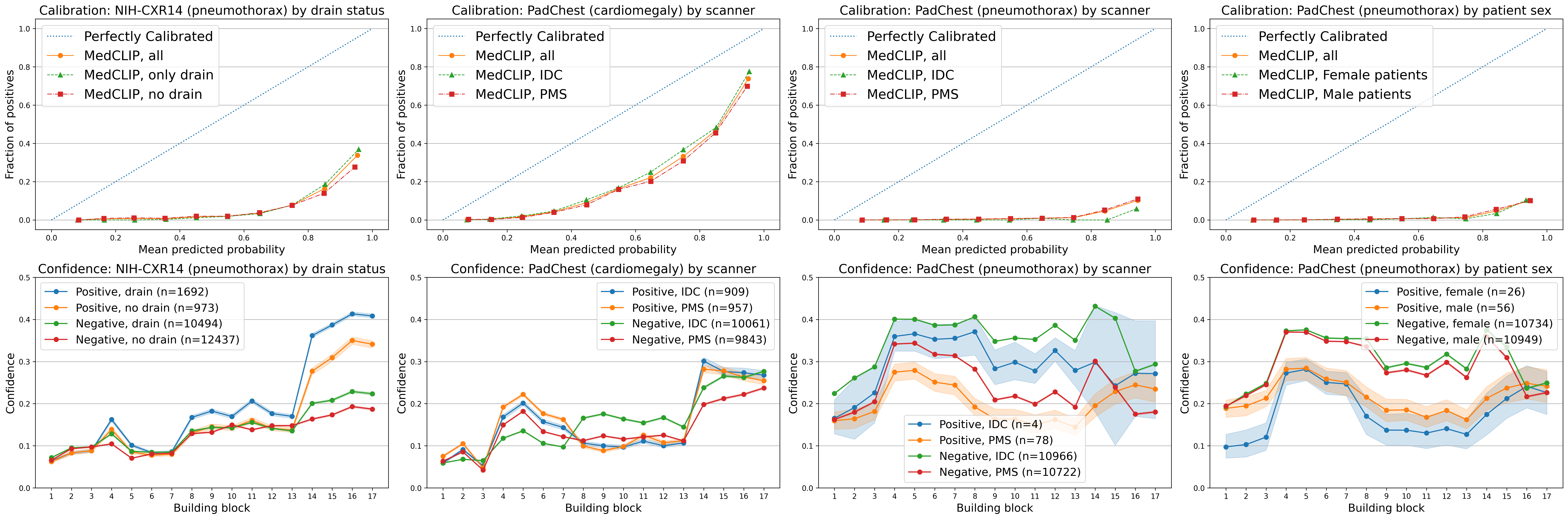}
    \caption{Calibration (top) and confidence (bottom) curves across all dataset configurations and their given subgroups.
    In the calibration curves, the x-axis shows the mean predicted probability per bin and the y-axis the fraction of positive cases. The diagonal represents perfect calibration, with deviations above or below the diagonal indicating miscalibration.
    In the confidence curves, the x-axis represents the individual building blocks with an attached probe, and the y-axis shows the mean confidence of the model's predictions, with 95\% confidence intervals obtained via $1,000$ bootstrap resamples per probe.}
    \label{fig:calibration_subfig}
\end{figure}

%% file: sec_09_lim_fut_con.tex
\input{figures/drain_examples}
Looking at the confidence throughout the building blocks of the vision encoder, we find signs of shortcut learning in the MedCLIP model, which aligns with previous research \cite{sourget2025fairness}. This suggests that, despite their scale and multi-modality, MedCLIP, and other CLIP-based models, are not inherently robust, even though they are state-of-the-art models. Furthermore, with these findings, we show evidence that the synthetic shortcuts presented by \cite{boland2025preventing} behave in a similar way as real-world shortcuts in medical datasets. We also find poor calibration in all models, aligning with the results of \cite{sourget2025fairness}. Our manual and exploratory data analysis uncovered several errors: for NIH-CXR14, we find cases of incorrectly entered patient ages and a greyed-out image. For PadChest, patients have conflicting sexes, duplicate images, duplicate metadata, as well as an X-ray of a skull. Further, we found the automatically annotated drains of the negative cases of pneumothorax in NIH-CXR14 were not completely reliable. From visually inspecting the images, we found cases where the images had been labelled with "no drain" but the images do actually contain a drain. This was the case for $3$/$5$ of the images we visually examine. These standout images that we found can be seen in \autoref{fig:drains_example}.
\\
\indent
Due to the data quality issues, the results should be interpreted carefully, as the conclusions are drawn from datasets that contain label and metadata errors.
Further, our results also highlight the limitations of methods such as calibration and confidence curves, which can be severely influenced by the number of positive cases available, such as in PadChest for pneumothorax classification. 
Moreover, as shortcuts are dataset-specific, experiments on other datasets may yield different findings. Similarly, we conducted our analysis on a single CLIP-based model and vision encoder. Conducting our experiments on other architectures and datasets would therefore help to improve the generalisation of our results. Finally, it would be beneficial to extend this analysis beyond CNN-based vision encoders to more recent architectures such as Transformers.
\\
\indent
To conclude, we applied the methodology of linear probes in the intermediate layers from \cite{boland2025preventing} to a new model architecture, namely MedCLIP, to study how real-world shortcuts emerge in the vision encoder across three different dataset configurations: NIH-CXR14 for pneumothorax, and PadChest for cardiomegaly and pneumothorax.
All dataset configurations showed miscalibrated models.
Some confidence curves revealed that the confidence did not develop in a stable manner across the network, but aligned with \cite{boland2025preventing}, who state that different shortcuts manifest at different depths of the network.
We revealed unreliable drain annotations for the negative-pneumothorax cases, and errors in the publicly available datasets NIH-CXR14 and PadChest. These findings further highlight that scale, architecture and high AUROC scores alone do not make SOTA models robust to shortcuts. The errors found underscore the need for high data quality to make reliable conclusions.

%% file: figures/drain_examples.tex
\begin{figure*}
    \centering
    \begin{subfigure}[t]{0.3\textwidth}
        \centering
        \includegraphics[width=\textwidth]{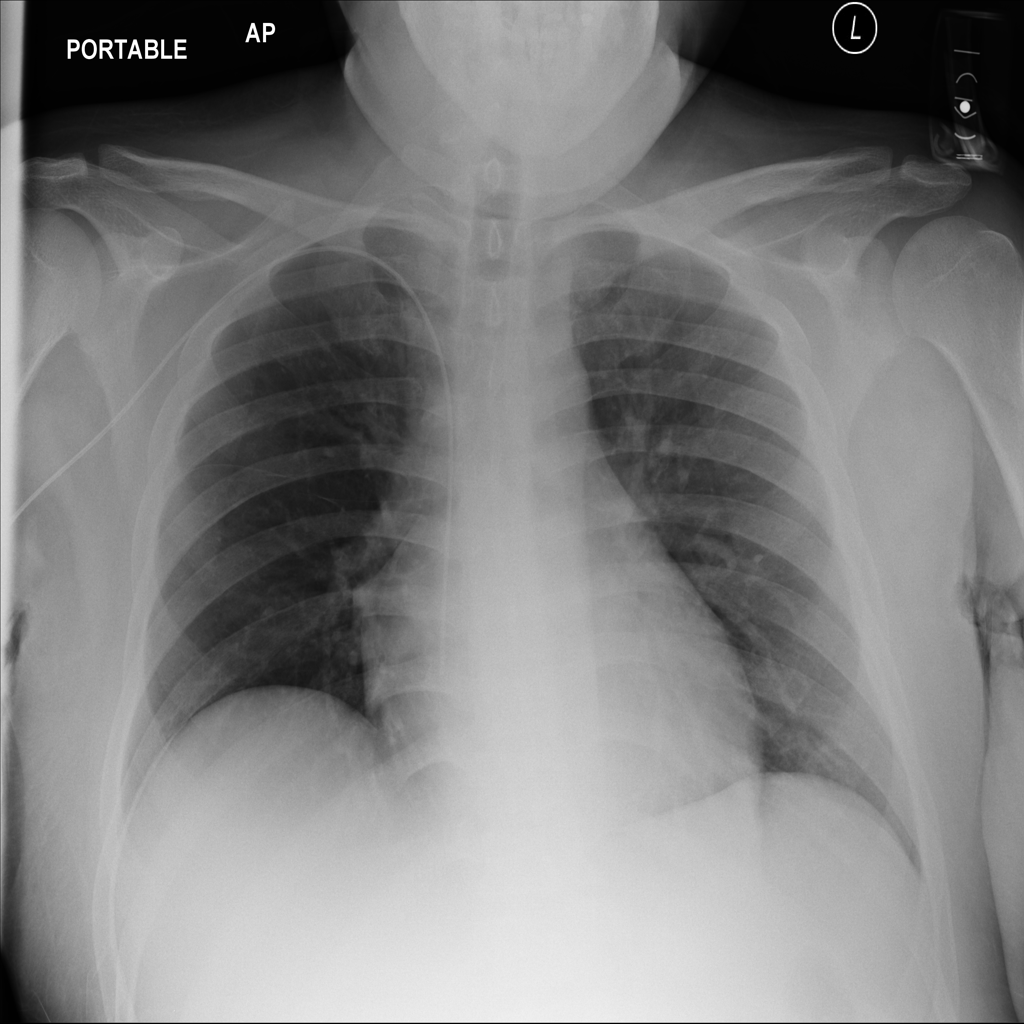}
        \caption{NIH-CXR14 (image ID 00020945\_039.png) has no pneumothorax. A small drain can be seen in the top left corner, but it was annotated as having no drain.}
        \label{fig:drain1}
    \end{subfigure}
    \hfill
    \begin{subfigure}[t]{0.3\textwidth}
        \centering
        \includegraphics[width=\textwidth]{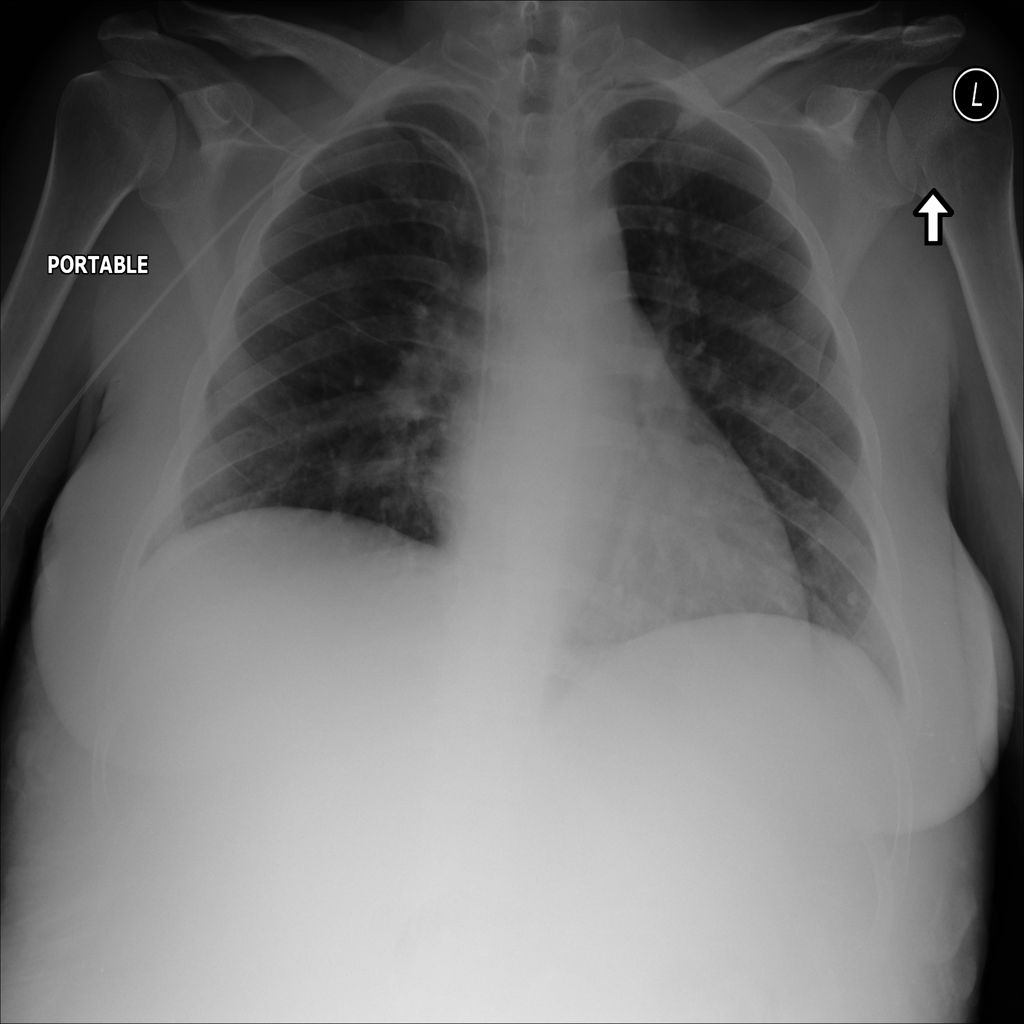}
        \caption{NIH-CXR14 (image ID 00013377\_010.png) has no pneumothorax. A small drain can be seen in the top left corner, but it was annotated as having no drain.}
        \label{fig:drain2}
    \end{subfigure}
    \hfill
    \begin{subfigure}[t]{0.3\textwidth}
        \centering
        \includegraphics[width=\textwidth]{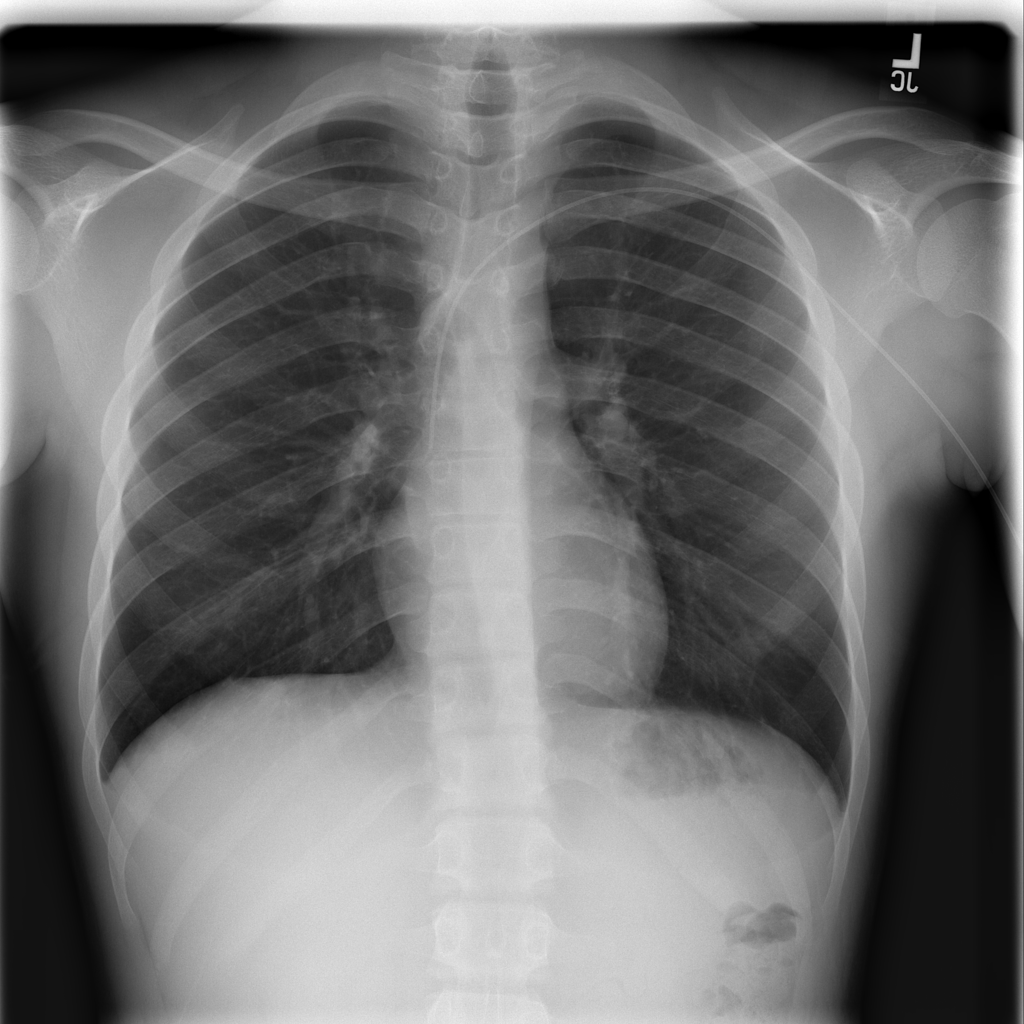}
        \caption{NIH-CXR14 (image ID 00017318\_015.png) has no pneumothorax. A small drain can be seen in the top right corner, but it was annotated as having no drain.}
        \label{fig:drain3}
    \end{subfigure}
    \par\vspace{1em}
    \begin{subfigure}[t]{0.3\textwidth}
        \centering
        \includegraphics[width=\textwidth]{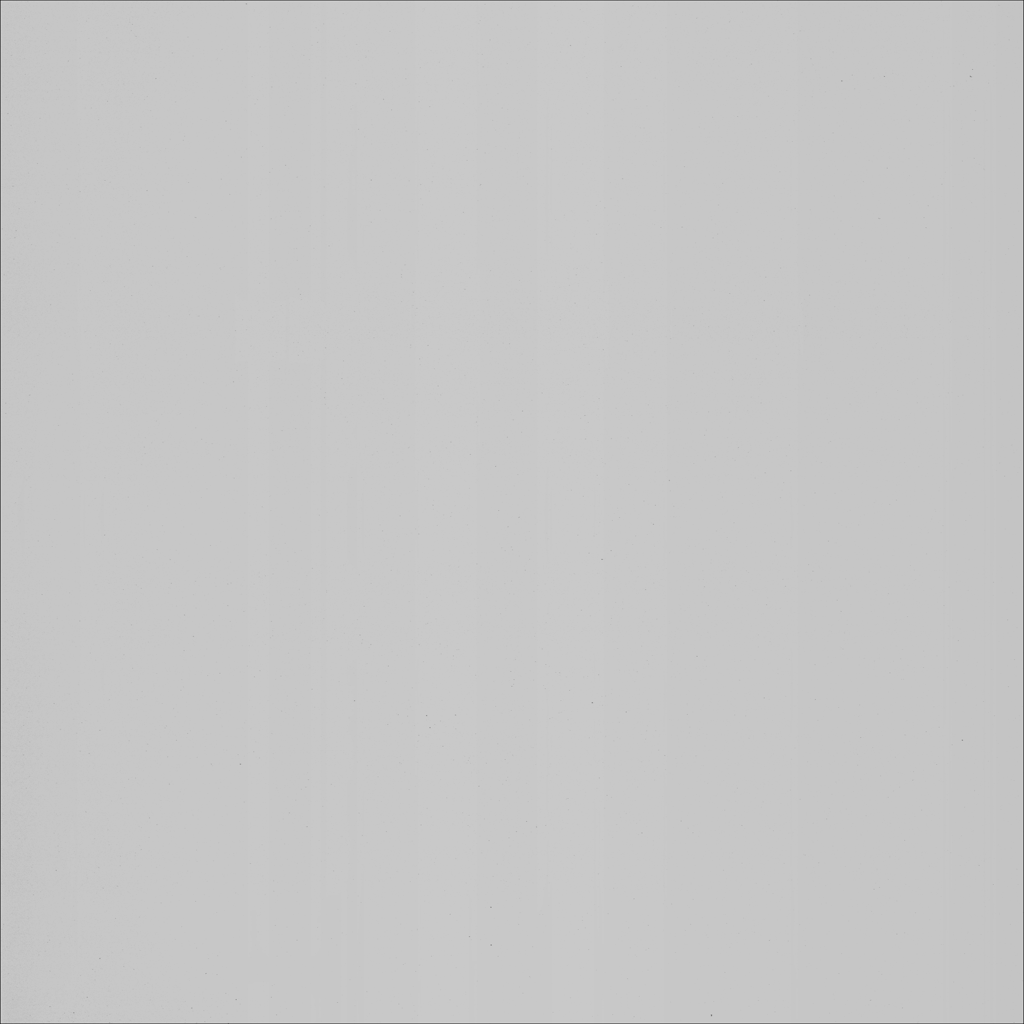}
        \caption{NIH-CXR14 (image ID 00010007\_121.png) is a non-uniform grey image.}
        \label{fig:grey}
    \end{subfigure}
    \hspace{0.05\textwidth}
    \begin{subfigure}[t]{0.3\textwidth}
        \centering
        \includegraphics[width=\textwidth]{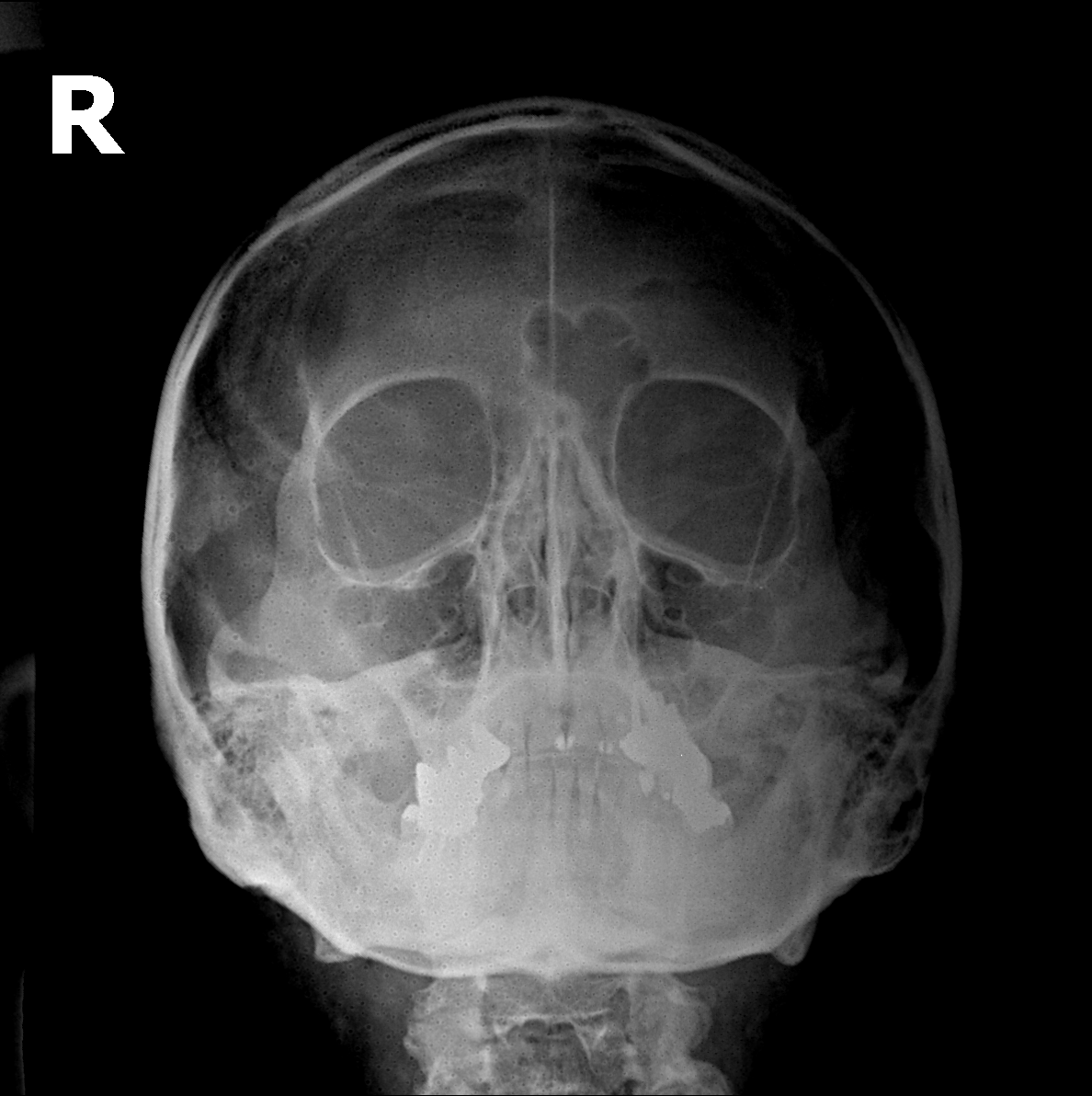}
        \caption{PadChest (image ID 216840111366964012339356563862\\
        009068132653048\_00-050-189.png) is not a chest X-ray, but an X-ray of a skull, annotated with lung diseases.}
        \label{fig:skull}
    \end{subfigure}
    \caption{Selected standout images identified during the manual analysis.}
    \label{fig:drains_example}
\end{figure*}

%% file: sec_10_ack.tex

\noindent VC and TS were supported by Novo Nordisk Foundation grant NNF24OC00926.

%% file: sec_11_interests.tex
The authors declare that they have no competing interests.